\documentclass[fleqn,10pt]{JLA_article}

\usepackage[english]{babel}
\usepackage{booktabs}
\usepackage{graphicx}
\usepackage{amsmath,amssymb}
\usepackage{array}
\usepackage{multirow}
\usepackage{caption}
\usepackage{url}
\usepackage{hyperref}

\makeatletter
\renewcommand{\Abstract}[1]{\long\def\@Abstract{#1}}
\providecommand{\abx@aux@refcontext}[1]{}
\providecommand{\abx@aux@cite}[2]{}
\providecommand{\abx@aux@segm}[3]{}
\providecommand{\abx@aux@defaultrefcontext}[3]{}
\providecommand{\abx@aux@defaultlabelprefix}[3]{}
\providecommand{\abx@aux@page}[2]{}
\providecommand{\abx@aux@fnpage}[2]{}
\renewcommand{\@maketitle}{%
\thispagestyle{fancy}%
{\raggedright\color{color1}\sffamily\bfseries\fontsize{20}{25}\selectfont \@PaperTitle\par}%
\vskip10pt%
{\raggedright\color{color1}\sffamily\fontsize{12}{16}\selectfont  \@Authors\par}%
\hfill\\
\sffamily\textbf{\abstractname}\\\@Abstract\\[4pt]%
\fcolorbox{color1}{white}{%
\parbox{\textwidth-2\fboxsep-2\fboxrule}{\centering%
\ifWithoutNotes
\parbox{\textwidth-6\fboxsep-2\fboxrule}{%
\textbf{Keywords}\\\@Keywords\\[4pt]%
\textbf{Submitted:} \@Submitted \hskip4pt ---  \textbf{Accepted:} \@Accepted  \hskip4pt ---  \textbf{Published:} \@Published
}
\else
\colorbox{color2!10}{%
\parbox{\textwidth-4\fboxsep-2\fboxrule}{%
\sffamily\textbf{\@Notesname}%
\begin{itemize}
\begingroup%
\@note%
\endgroup
\end{itemize}
}%
}%
\vskip4pt%
\parbox{\textwidth-4\fboxsep-2\fboxrule}{%
\textbf{Keywords}\\\@Keywords\\[4pt]%
\textbf{Submitted:} \@Submitted \hskip4pt ---  \textbf{Accepted:} \@Accepted  \hskip4pt ---  				\textbf{Published:} \@Published
}
\fi
}%
}%
\\
\begingroup%
\raggedright\sffamily\small%
\footnotesize\@affiliation\par%
\endgroup
\vskip25pt%
}
\makeatother

\hypersetup{hidelinks,colorlinks,breaklinks=true,urlcolor=color2,citecolor=color1,linkcolor=color1,bookmarksopen=false,pdftitle={Study-Strategy Clusters from EdNet Logs Track Engagement, Not Mastery},pdfauthor={Qingchuan Lyu, Yingxin Li, Albert Yang}}

\graphicspath{{project_graphs/}{./}}

\newcommand{\Nusers}{5{,}000}
\newcommand{\Nfeatures}{4{,}999} 
\newcommand{\Ndropped}{13} 
\newcommand{\Nclustered}{4{,}986}
\newcommand{\Neligible}{45{,}621}
\newcommand{\StabARImin}{0.965} 
\newcommand{\StabNrep}{3} 
\newcommand{\LockedK}{5}
\newcommand{\Npoles}{4}
\newcommand{\Nhier}{8}
\newcommand{\MajorityPct}{{$\sim$64.9\%}}
\newcommand{\MajSubN}{3{,}236}
\newcommand{\MajSubK}{4}
\newcommand{\MajSubSil}{0.199}
\newcommand{\FlatSil}{0.333}
\newcommand{\StabARImean}{0.984}
\newcommand{\HierStabARImean}{0.938}
\newcommand{\PolesStabARImean}{0.986}
\newcommand{\MajSubStabARImean}{0.915}
\newcommand{\SeqARIbasemean}{0.986}
\newcommand{\SeqARImean}{0.686}
\newcommand{\StabARI}{{mean ARI $\approx \StabARImean$}}
\newcommand{\HierStabARI}{{mean ARI $\approx \HierStabARImean$}}
\newcommand{\NclustFeatures}{9}
\newcommand{\NPCA}{7}
\newcommand{\PcaVar}{97.8\%}
\newcommand{\WinsorCap}{1}
\newcommand{\ValEarlyN}{4{,}960}
\newcommand{\ValARI}{0.253}
\newcommand{\ValEarlyPersistEta}{0.106}
\newcommand{\ValEarlyCompleteEta}{0.021}
\newcommand{\ValEarlyTermEta}{0.003}
\newcommand{\ValEarlyTermPadj}{0.093}
\newcommand{\OutScorable}{2{,}952}
\newcommand{\OutScorableLabeled}{2{,}939}
\newcommand{\VolARISameK}{0.064}
\newcommand{\VolRespondEta}{0.095}
\newcommand{\VolMasteryEta}{0.001}
\newcommand{\StratMasteryEta}{0.003}
\newcommand{\MasteryOmnibusPP}{10.48}
\newcommand{\KTauc}{0.605}
\newcommand{\KTbase}{0.553}
\newcommand{\KTlift}{{$+0.051$}}
\newcommand{\KTliftCI}{{$[+0.045,+0.058]$}}
\newcommand{\KTnUsers}{4{,}994}
\newcommand{\KTnSkills}{7}
\newcommand{\KTvsBehARI}{0.007}
\newcommand{\KTtermEta}{0.017}
\newcommand{\KTpcOner}{{$+0.785$}}
\newcommand{\KTpcOnevar}{0.380}
\newcommand{\KTabilityR}{0.252}
\newcommand{\DiffVideoBPP}{{$-12.7$}}
\newcommand{\DiffVideoBootCI}{{$[-20.9,-4.2]$}}

\newcommand{\MajSubStabARI}{{mean ARI $\approx \MajSubStabARImean$}}
\newcommand{\MajSubStabARImin}{0.886}

\newcommand{\SeqReviserPct}{{$\sim$71\%}}
\newcommand{\SeqAnswerPct}{{$\sim$29\%}}
\newcommand{\SeqPersistEta}{0.038}
\newcommand{\SeqTermEta}{0.001}

\PaperTitle{Study-Strategy Clusters from EdNet Logs Track Engagement, Not Mastery}

\Authors{Qingchuan Lyu\textsuperscript{1*}, Yingxin Li\textsuperscript{2}, Albert Yang\textsuperscript{1}}
\affiliation{\textsuperscript{1}\textit{Georgia Institute of Technology, Atlanta, GA, USA}}
\affiliation{\textsuperscript{2}\textit{New York University, New York, NY, USA}}
\affiliation{\textsuperscript{*}\textit{Corresponding author:}~\texttt{qlyu9@gatech.edu}}

\Keywords{learning analytics, learner profiling, clustering, engagement, mastery, knowledge tracing, intelligent tutoring systems}

\Submitted{18/08/2026}
\Accepted{--}
\Published{--}

\Notesname{Notes for Practice}

\note{Established knowledge: learning analytics and Intelligent Tutoring System (ITS) research often cluster interaction logs, such as resource use, revision, video watching and problem attempts, into named learner types or study styles. Internal validation metrics (silhouette, Davies--Bouldin) and clear visual separation in embeddings are then frequently taken as evidence that those styles are educationally meaningful, including as proxies for how well learners know the material, without a separate check against later correctness or external assessments.}
\note{Contribution of this paper: on EdNet-KT3 we recover a stable hierarchy of study-strategy clusters from early practice logs. To test whether those clusters forecast learning, we split each learner's timeline in half by respond count so cluster features use only the early half and outcomes use only the late half (blocking leakage from late behavior into both sides). Early clusters predict later engagement---how much learners keep practicing and finish sessions in that late half, especially staying active across late sessions---but not later unassisted accuracy (correctness on late first-attempts answered without help). Checks on practice volume, answer-order patterns, and a knowledge-tracing model still find no reliable mastery signal from behavior-only styles.}
\note{Implications for practice: use log-based clusters to describe study styles and to target engagement support (e.g., persistence, completion), not as standalone predictors of knowledge gains; if mastery is the goal, prefer correctness-aware models and external outcome measures rather than behavior-only labels.}

\Abstract{Learning analytics often treats unsupervised clusters of intelligent tutoring system (ITS) logs as learner types that should predict learning. We test that assumption on EdNet-KT3. Clustering study-strategy features (resource use, revision, video, problem practice) for \Nusers\ active learners yields a silhouette-selected parent cut ($k=\LockedK$) with \Npoles\ contrast poles (reading-focused, video-heavy, revision-heavy, and problem-first) plus a large near-mean residual (\MajorityPct). Reclustering that residual adds four finer styles, giving a bootstrap-stable hierarchy of \Nhier\ named strategies. We split each learner's timeline by respond count so clusters use only the early half and outcomes only the late half. Early clusters predict later engagement (continuing to practice and finishing late sessions, especially persistence, $\eta^{2}\approx\ValEarlyPersistEta$; completion $\eta^{2}\approx\ValEarlyCompleteEta$) but not later unassisted accuracy (correctness on late first-attempts without help; $p_{\mathrm{adj}}\approx\ValEarlyTermPadj$). Volume rises with some styles, yet volume-only clustering barely matches strategy labels (ARI$=\VolARISameK$). A knowledge-tracing model (SAKT) on the seven TOEIC exam sections predicts next correctness only modestly better than a baseline that knows only how hard each section usually is (AUC lift \KTlift; CI \KTliftCI), and that mastery signal is nearly independent of behavior styles (ARI$=\KTvsBehARI$). Behavioral clustering here describes study styles and engagement, not knowledge gains.}

\begin{document}

\flushbottom
\maketitle
\thispagestyle{fancy}

\section{Introduction}
\addcontentsline{toc}{section}{Introduction}

Intelligent tutoring systems (ITS) log everyday study moves, such as watching video,
reading explanations, answering and revising questions. A common hope in
learning analytics is that clustering those logs will reveal learner types that
\emph{predict} learning outcomes and can drive adaptive support
\cite{baker2009jedm,desmarais2012review}. That hope is rarely stress-tested:
many pipelines stop once clusters look interpretable and well separated
(silhouette scores, UMAP plots), treating educational meaning as self-evident
\cite{vendramin2010comparative}.

\textbf{Research question.} Do unsupervised study-strategy profiles from ITS
logs forecast later \emph{learning}, or only later \emph{engagement} (whether
learners keep practicing)? We answer this on the public EdNet-KT3 corpus
\cite{choi2020ednet}. First we discover and name styles on each learner's full
log. Then, for prediction claims, we rebuild clusters from only the early half
of each learner's question attempts and score outcomes on the late half. This way
late behavior cannot leak into both the cluster features and the scored
outcome.

Our contribution is threefold. (i)~We recover a stable hierarchy of study
styles: \Npoles\ sharp contrast poles (reading focused, video heavy, revision heavy, and
problem attempt first) plus \MajSubK\ softer styles inside the large near-average
majority (mild in-session study before attempting problems, video-leaning majority, read passages then attempt, mild revisers). Parent $k{=}\LockedK$ and the hierarchy stay stable under bootstrap
resampling (\StabARI; \HierStabARI). (ii)~When clusters are fit on early
practice only, they predict later engagement, especially whether learners stay
active across late sessions, but not correctness on
late first-attempts answered without help. (iii)~Checks on practice volume,
answer-order patterns, and a supervised knowledge-tracing model (SAKT)
\cite{pandey2019sakt} do not overturn that mastery null: SAKT can separate
mastery, but as an ability gradient nearly independent of behavior-style labels.
For practice, log-based styles are useful for describing \emph{how} learners
study and whether they keep practicing, not as a proxy for \emph{what} they
know.

\section{Literature Review}

This review situates the paper at the intersection of intelligent tutoring
systems (ITS), learning-style and learning-strategy personalization,
self-regulated learning (SRL), log-based behavioral profiling, and the
distinction between engagement and learning outcomes. Prior work often
personalizes to named learning styles or predicts next-item correctness; far
less work asks whether \emph{unsupervised behavioral clusters} derived from
study-strategy features forecast later mastery when early cluster features are
kept separate from late outcomes, or whether the same clusters that predict
engagement also predict achievement. That gap is our target.

\subsection{Intelligent tutoring and adaptive instruction}
Cognitive tutors established that model-tracing ITS can improve classroom
outcomes when they provide immediate, concise feedback grounded in a cognitive
model of the domain \cite{anderson1995cognitive,koedinger1997intelligent}.
Later surveys framed open challenges in student modeling, metacognition, and
authoring \cite{conati2009intelligent,murray2003overview}, and extended tutor
principles to help-seeking and self-explanation
\cite{aleven2002effective,roll2007designing,azevedo2002beyond}. Adaptive
hypermedia and personalized ITS emphasize individual goals, preferences, and
prior knowledge as bases for adaptation
\cite{brusilovsky2000adaptive,phobun2010adaptive}.

Systematic reviews catalog AI techniques used in ITS (rule-based methods,
Bayesian networks, and data mining) and stress evaluation on learner outcomes
\cite{mousavinasab2021intelligent,lin2023artificial,Hariyanto_Kristianingsih_Maharani_2025,Gligorea2023-rb}.
A 2025 systematic review of 28 K--12 studies ($N{=}4{,}597$) found ITS effects
generally positive yet often mitigated when compared with non-intelligent
tutoring systems, highlighting personalization and adaptivity as critical
design levers \cite{letourneau2025its}. Implementation studies further show that
adaptive tutors succeed or fail as much on deployment and instructional fit as
on the underlying learner model
\cite{phillips2020implementing,erumit2020design}. Collectively, this work
underscores that \emph{how} students engage with an ITS, the strategies and
patterns they deploy, may matter as much as the system's algorithm.

We inherit the ITS goal of actionable learner models, but ask a narrower
question: whether \emph{unsupervised} study-strategy clusters from interaction
logs forecast later mastery, or mainly later engagement.

\subsection{Learning styles and log-based profiling}
A large adaptive-learning literature identifies learning styles to personalize
instruction \cite{Essa2023-xt,Bajaj2018-py}. The Felder--Silverman taxonomy
remains influential; methods range from questionnaires to hybrid neural style
recognizers \cite{Bernard2017-cx,Bernard2022-jn} and clustering of web-usage
traces under Felder--Silverman dimensions \cite{Azzi2020-xp}. Reviews of
technology-enhanced personalized learning likewise highlight data-driven
adaptation as a central trend \cite{Xie2007-nd}. Unsupervised clustering of
log-derived features is also a long-standing approach to discovering study
strategies and learner types in educational data mining and learning analytics
\cite{baker2009jedm,desmarais2012review}. Typical pipelines aggregate action
counts or time-on-task, reduce dimensionality, and interpret clusters as
profiles for adaptation.

Our approach is related but distinct: we cluster simple study-strategy
summaries---how often learners watch video, read, attempt, revise, study before
answering, and return after errors---rather than map learners onto a fixed
questionnaire taxonomy, and we judge the clusters by both later engagement and
later mastery. Silhouette and similar scores help choose how many clusters to
keep, but they do not prove educational meaning
\cite{vendramin2010comparative}; we therefore also require that labels stay
stable under resampling \cite{vonluxburg2010clustering,hubert1985ari} and that
prediction claims use held-out late outcomes.

\subsection{Self-regulated learning and metacognition in ITS}
Self-regulated learning (SRL) provides a theoretical backbone for how learners
deploy strategies in tutoring systems. Winne and Hadwin's COPES model describes
SRL across recursive phases: task definition, goal setting and planning,
studying tactics, and adaptations, with metacognitive monitoring as a gateway
to regulation \cite{winne1998self}. Prompting self-explanation in cognitive
tutors, a lightweight metacognitive scaffold, produced greater learning gains
and transfer than procedural practice alone \cite{aleven2002effective}. Related
work shows that students often deploy ineffective help-seeking strategies, and
that metacognitive tutoring can improve both help-seeking and learning
\cite{roll2007designing,azevedo2002beyond}.

Furthermore, a 2025 systematic mapping of 84 AI--SRL studies found that AI is most often
used for adaptive personalization, prediction/profiling, ITS, and assessment,
with metacognitive and cognitive aspects far more studied than motivation
\cite{banihashem2025aisrl}. That imbalance matters for log-based profiling:
behavior traces readily capture strategy and engagement signals, but motivational
constructs are harder to recover from actions alone. We treat our study-strategy
clusters as descriptive behavioral encodings of how learners regulate practice
in the log, not as full SRL diagnoses.

\subsection{Engagement versus achievement}
Engagement (participation, persistence, time-on-task) is often treated as a
proxy for learning, yet the two can come apart: learners can be highly active
without commensurate gains, and vice versa
\cite{fredricks2004school,henrie2015measuring}. Recent AI-tutor and GenAI
education work similarly cautions that gains in interaction efficiency,
engagement, or process quality need not yield immediate learning gains
\cite{kasneci2023chatgpt,francis2025generative,Kong2025-ta}. We therefore score
both engagement (whether late sessions continue and items are completed) and
mastery (accuracy on late first-attempts answered without help), so a cluster
that tracks who keeps practicing is not mistaken for a cluster that tracks who
knows more. This distinction matters in practice: an intervention keyed to
``revision-heavy'' membership may find who stays active later without diagnosing
who knows more.

\subsection{Knowledge tracing and outcome prediction}
Predicting next-response correctness from interaction sequences is a core EDM
task. Classical Bayesian Knowledge Tracing (BKT) \cite{corbett1995bkt} models
per-skill binary latent knowledge with a Hidden Markov update and remains a
foundational baseline. Self-Attentive Knowledge Tracing (SAKT)
\cite{pandey2019sakt} and later transformer-style models are now standard neural
baselines on large public corpora; related surveys cover dynamic KT for
large-scale adaptive environments \cite{Sapountzi2019-wg} and continuous
personalized KT \cite{Wang2023-na}. EdNet \cite{choi2020ednet} provides
multi-million-interaction TOEIC tutoring logs; each question carries a
\texttt{part} field taking values $1$--$7$ (Listening \& Reading exam sections)
plus finer expert \texttt{tags}. Published full-corpus KT benchmarks often
report AUC near $\sim$0.75--0.78.

We use EdNet-KT3 not to chase that leaderboard, but as a realistic ITS log for
testing whether \emph{behavioral} structure predicts later in-app mastery. Our
supervised probe uses SAKT to predict next-response correctness on the coarse
\texttt{part} labels. Because some exam sections are simply harder than others,
a strong baseline is a \emph{part-difficulty prior}: for an item in section~$p$,
guess correctness using only how often the cohort gets section~$p$ right---the
same guess for every learner, with no personal answer history. We judge SAKT by
AUC lift over that baseline, not by absolute state-of-the-art AUC, and
critically we also cluster SAKT's dense student embedding as a
correctness-supervised contrast to unsupervised behavioral clusters.

\subsection{Personalization in the era of large language models}
The mid-2020s convergence of ITS ideas with large language models creates new
personalization opportunities and new risks. Reviews and commentaries emphasize
both promise (dialogic tutoring, scalable feedback) and limits (uneven
pedagogical quality, equity, assessment integrity)
\cite{kasneci2023chatgpt,abd2023large,gabriel2024generative,francis2025generative,abedin2026pedagogical,garcia2026perspective}.
Recent adaptive and predictive tutoring systems continue to combine mastery
estimates, engagement signals, and personalized guidance
\cite{Duque2026-bd,Kong2025-ta,Hariyanto_Kristianingsih_Maharani_2025}. A
recurring theme is that personalization may improve the \emph{process} of
learning (engagement, cognitive load, efficiency) more reliably than immediate
\emph{outcomes}, which is echoing the engagement--achievement dissociation above. Careful
integration with traditional learner modeling, rather than wholesale
replacement, remains the more promising path for claims about mastery.

\subsection{Hierarchy, stability, and predictive validity}
Methodological work on cluster validation warns that high silhouette or a clean
UMAP plot does not imply external validity
\cite{vonluxburg2010clustering,hubert1985ari,vendramin2010comparative}.
Education data-mining practice often reports a chosen $k$ and named profiles;
fewer papers publish bootstrap ARI floors, feature-definition ablations, and
temporally held-out outcome tests in the same manuscript. Hierarchical
decompositions of a large residual majority are also under-reported relative to
flat $k$-means cuts. We treat the \Npoles\ poles + \MajSubK\ majority-style
hierarchy as a descriptive encoding whose predictive claims must be earned
separately under anti-leakage evaluation, with temporal holdout, feature
ablation, and bootstrap stability accompanying any prediction claim.

\subsection{Formal background}
Three mathematical ingredients frame our design.
\emph{(i)~Ability vs.\ behavior.} Item-response theory (IRT) models correctness
as a function of latent ability \(\theta\) and item difficulty \(b\). In the
one-parameter logistic (Rasch) form \cite{rasch1960,embretson2000irt},
\begin{equation}
  P(X_{ij}{=}1\mid \theta_i,b_j)
  = \frac{1}{1+\exp\!\bigl(-(\theta_i-b_j)\bigr)}.
  \label{eq:irt}
\end{equation}
Models trained on correctness (including neural KT latents) are therefore
expected to recover an ability-like direction; clustering such latents can
separate mastery without implying a \emph{behavioral} typology.
\emph{(ii)~Order as a Markov process.} Sequence profiles treat within-session
actions as a first-order chain with transition matrix \(P(a'{|}a)\). Aggregate
(``flat'') features discard \(P(a'{|}a)\) and keep only frequency of action and
time share of actions; exploratory order probes ask whether residual structure lives in
that transition matrix.
\emph{(iii)~Beyond silhouette alone: Stability.} Partition quality under resampling is
summarized by the chance-corrected adjusted Rand index (ARI)
\cite{hubert1985ari}:
\begin{equation}
  \mathrm{ARI}
  = \frac{\mathrm{RI}-\mathbb{E}[\mathrm{RI}]}
         {\max(\mathrm{RI})-\mathbb{E}[\mathrm{RI}]},
  \label{eq:ari}
\end{equation}
so \(\mathrm{ARI}{\approx}1\) indicates reproducible structure beyond chance
\cite{vonluxburg2010clustering}. We gate on bootstrap ARI, then test outcomes
with ANOVA, Holm correction, and bootstrap confidence intervals on mastery gaps
\cite{efron1993bootstrap}.

\subsection{Summary of the gap}
The literature shows that ITS can support learning when feedback and adaptive
modeling are well designed; that learning-style and log-based profiles are
widely used for personalization; that SRL theory explains why strategy and
metacognition matter; that engagement and achievement can diverge; and that KT
and LLM tutors improve prediction or process without automatically validating
behavior-only mastery claims. What remains underexplored is a unified empirical
treatment that (1)~derives interpretable study-strategy clusters from ITS logs,
(2)~evaluates predictive validity separately for engagement and mastery under
strict temporal holdout, (3)~compares those unsupervised clusters against a
supervised correctness embedding (SAKT), and (4)~validates stability via
bootstrap resampling and feature ablation. That is the target of the present
work.

\section{Methods}

\subsection{Cohort, vocabulary, windowing and features}
EdNet-KT3 provides one timestamped action log per user ($\sim$298{,}000 users
scanned). We define an \emph{active learner} as a user with at least 50 question
\texttt{respond} events (not total actions or calendar days). Under this gate,
\Neligible\ users were eligible. Profile discovery, outcome audits, and the KT
probe use a fixed seeded sample of \Nusers\ learners; \Nclustered\ remain after
dropping rows with non-finite clustering features (typically division-by-zero when a user has no timed study; Table~\ref{tab:cluster_definitions}). The activity threshold ensures enough
interactions for session windowing, but introduces activity bias: results
describe engaged users, not the full EdNet population. 

\begin{table}[ht]
\caption{Locked clustering features (\NclustFeatures), grouped by flat
study-strategy category. Action-mix values are shares of non-idle actions. A
feature is non-finite (and the row is dropped before PCA) when its denominator
is zero---almost always \texttt{dwell\_frac\_watch\_video} with no timed study
($0/0$); rarely also \texttt{study\_before\_answer\_ratio},
\texttt{revision\_rate}, or \texttt{reengage\_after\_error\_rate} when the
corresponding count is zero.}
\label{tab:cluster_definitions}
\centering
\small
\setlength{\tabcolsep}{4pt}
\begingroup
\renewcommand{\_}{\textunderscore\allowbreak}%
\begin{tabular}{@{}
  >{\raggedright\arraybackslash\small}p{0.18\linewidth}
  >{\raggedright\arraybackslash\ttfamily\footnotesize}p{0.30\linewidth}
  >{\raggedright\arraybackslash\small}p{0.44\linewidth}
  @{}}
\toprule
\normalfont Category &
\multicolumn{1}{l}{\normalfont Feature} &
\normalfont Definition \\
\midrule
\multirow{5}{=}{Action-mix frequencies}
  & frac\_n\_watch\_video
  & Frequency of non-idle actions that are video \\
  & frac\_n\_read\_explanation
  & Frequency that are explanation reads \\
  & frac\_n\_read\_passage 
  & Frequency that are passage / stem reads \\
  & frac\_n\_attempt\_problem
  & Frequency that are first attempts \\
  & frac\_n\_revise\_answer
  & Frequency that are revises \\
\cmidrule(l){1-3}
\multirow{1}{=}{Video dwell share}
  & dwell\_frac\_watch\_video
  & Timed study seconds on video / total timed study seconds.
    Passage dwell is constant zero on KT3 (excluded); explanation dwell is
    near-complementary and dropped to avoid collinearity. \\
\cmidrule(l){1-3}
\multirow{3}{=}{Strategy rates}
  & study\_before\_answer\_ratio
  & First attempts preceded by in-session video or explanation divided by
    all first attempts \\
  & revision\_rate
  & Revises / attempts (winsorized at \WinsorCap) \\
  & reengage\_after\_error\_rate
  & Later returns to a wrong item / wrong responds \\
\bottomrule
\end{tabular}
\endgroup
\end{table}

Actions are mapped to a compact vocabulary:
watch\_video, read\_passage, read\_explanation,
attempt\_problem, revise\_answer, and idle\_return.
An \texttt{attempt\_problem} is the first response to a question; a later
response to the same question is a \texttt{revise\_answer}. Streams are
segmented with a 30-minute inactivity gap. Sessions with fewer than five
actions are dropped. Each user's log is capped at the earliest $5{,}000$
actions so extreme histories, such as repetitive revising answers many times, do not dominate computation or overweight
late-career drift. Session windows roll up to user-level features deliberately disjoint from outcomes. The locked clustering set uses \NclustFeatures\ features
(Table~\ref{tab:cluster_definitions}). Features are standardized and reduced with PCA (\NPCA\ components retaining
\PcaVar\ of variance). K-means clustering is performed in this PCA space; UMAP
is used only for visualization.

\subsection{Hierarchy, $k$ selection, and stability}
We choose the number of parent clusters by silhouette and Davies--Bouldin ranks over $k{\in}[2,12]$,
which selects $k{=}\LockedK$ (silhouette $\approx$\FlatSil;
Figure~\ref{fig:select-k}). The largest cluster is still a near-average residual
(\MajorityPct; \MajSubN\ users), so we run a second K-means only on that group
and select $k{=}\MajSubK$ by combined silhouette and Davies--Bouldin ranks
(majority silhouette $\approx$\MajSubSil). The reported typology is therefore
hierarchical: \Npoles\ contrast poles kept as parent labels plus
\MajSubK\ majority styles (\Nhier\ named learning patterns in total).

To check stability, we re-cluster independent 80\% subsamples at fixed $k$ and
measure agreement with full-cohort labels using the adjusted Rand index (ARI)
\cite{hubert1985ari,efron1993bootstrap}. Density-based HDBSCAN on the same PCA
space labels everyone as noise at the parent stage, so we do not use it for
profile definition. Feature-definition ablations (merged reading categories,
longer sessions, mix/dwell/strategy-only, log-counts) show that maximizing
silhouette alone can favor coarser partitions; we retain the locked hierarchy
for interpretability and stability, \textit{not} maximum silhouette.

\subsection{Anti-leakage outcome design}
Table~\ref{tab:sample-sizes} lists recurring sample sizes so changing $n$
values are clear in various situations. Profile discovery uses the clustered cohort; mastery tables further require finite \texttt{terminal\_unassisted\_accuracy} (last 20
late unassisted first-attempts, not merely ``$\geq$20 late responds,'' since
all \Nusers\ already meet a $\geq$5 late-respond gate).

\begin{table}[ht]
\caption{Recurring user counts on the locked analysis path.}
\label{tab:sample-sizes}
\centering
\small
\begin{tabular}{lrp{0.46\linewidth}}
\toprule
Inclusion rule & Users Count ($n$) & Role in analysis \\
\midrule
Active eligible ($\geq$50 responds) & \Neligible & Sampling frame before seeding \\
Seeded manifest & \Nusers & Locked analysis cohort \\
Finite clustering features & \Nclustered & Parent / hierarchy labeling \\
In typical-majority parent & \MajSubN & Second-stage majority styles \\
Finite terminal accuracy & \OutScorable & Mastery proxy available \\
Labeled $\cap$ scorable & \OutScorableLabeled & Mastery and volume--mastery tables \\
\texttt{strict\_early} clustered & \ValEarlyN & Early-half outcome tests \\
KT sequences (SAKT) & \KTnUsers & Knowledge-tracing probe
  ($\geq$10 first-attempts on \texttt{part} KCs) \\
\bottomrule
\end{tabular}
\end{table}

If we cluster on a learner's full log and then ``predict'' late accuracy from
those clusters, late practice can enter both the cluster features and the
outcome---an unfair circular test (leakage). We therefore split each learner's
question attempts in half by count (default 50/50 respond split): rebuild
features from the early half only, refit scaler/PCA/K-means at locked
$k{=}\LockedK$ (\texttt{strict\_early}; \ValEarlyN\ users), and score outcomes
only on the late half. Fixed $k$ is intentional: we ask whether the same number
of styles still forecasts late outcomes from early data, not whether early data
would choose a different $k$. Named poles and majority styles remain
full-timeline descriptions; early cluster ids are a fresh partition and should
not be narrated with those names.

Why split by respond count rather than calendar date? EdNet-KT3 users start and
stop at different times, and total practice volumes differ by orders of
magnitude; equalizing experience \emph{share} is the natural control. The cost
is unequal late \emph{volume}: some learners have few late items, so their
accuracy estimates are noisier. Users need at least 5 late responds; our mastery
proxy is accuracy on the last 20 late first-attempts answered without help
(\OutScorable\ scorable of \Nusers). Engagement proxies include whether late
sessions continue (\texttt{persistence\_late\_sessions}) and whether late items
are completed (\texttt{late\_completion\_rate}). We test cluster--outcome
differences with one-way ANOVA, Holm-adjusted $p$-values within each design, and
effect sizes ($\eta^2$).

\subsection{Volume audit and supervised KT probe}
To test whether parent styles are merely ``high vs.\ low practice volume''
groups, we compare average responds, actions, and sessions across parents, and
separately re-cluster the same users on those three volume features alone at
$k{=}\LockedK$. Separately, SAKT ($d{=}64$, 4 heads, max length 100, dropout
$0.2$, Adam $10^{-3}$, early stopping on validation AUC) predicts
next-response correctness on first-attempt sequences. We label items by the
\KTnSkills\ TOEIC \texttt{part} sections ($1$--$7$), not by EdNet's finer skill
\texttt{tags}: the goal is a correctness-supervised contrast to behavior
clusters (aligned with a section-difficulty baseline), not maximum KT accuracy.
The probe uses \KTnUsers\ users with $\geq$10 scorable responds. As a simple
baseline on the \emph{same} held-out attempts, we predict correctness from
section difficulty alone: if the next item is in TOEIC part~$p$, use the
cohort's average accuracy on part~$p$, and ignore who the learner is and how
they answered earlier items. SAKT is judged by how much its AUC beats that
baseline (a user-level bootstrap 95\% CI on
$\mathrm{AUC}(\mathrm{SAKT})-\mathrm{AUC}(\mathrm{prior})$). We also cluster
each user's mean SAKT hidden state ($d{=}64$) as a correctness-supervised
contrast to the behavior-only typology.

\section{Results}

\subsection{A stable hierarchical typology}
\begin{figure}[ht]
\centering
\begin{minipage}[t]{0.45\linewidth}
\centering
\includegraphics[width=\linewidth]{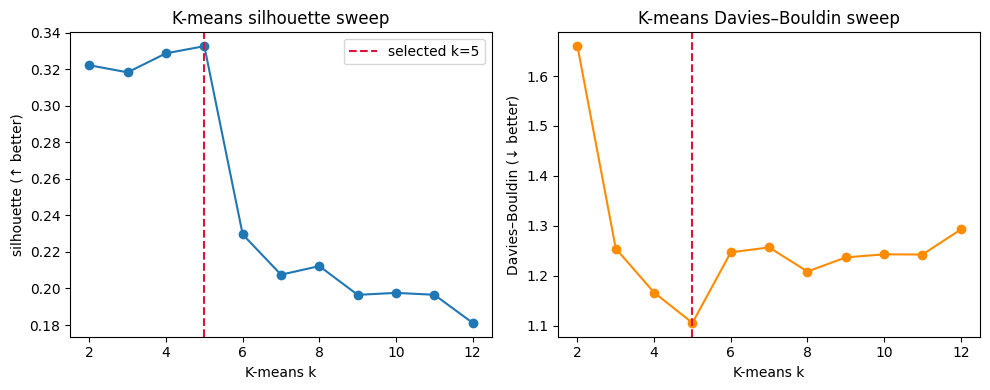}
\end{minipage}\hfill
\begin{minipage}[t]{0.48\linewidth}
\centering
\includegraphics[width=\linewidth]{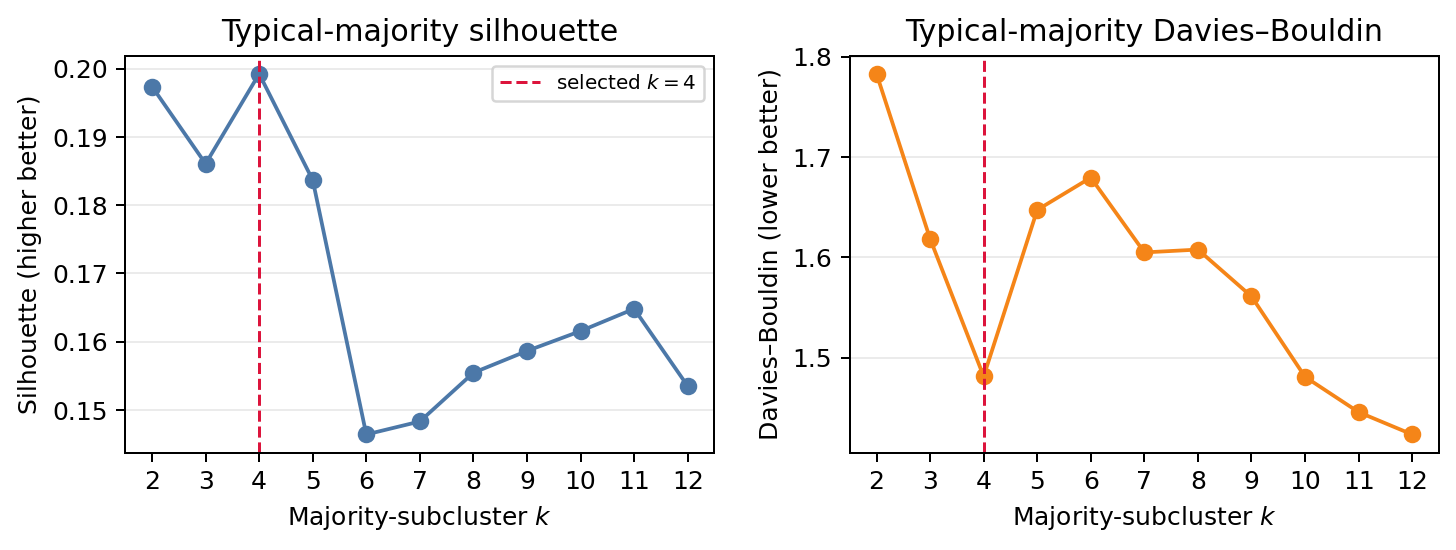}
\end{minipage}
\caption{Internal-validity sweeps for $k{\in}[2,12]$.
Left: parent-stage K-means; silhouette and Davies--Bouldin selects $k{=}\LockedK$ (dashed).
Right: typical-majority recluster; combined silhouette and Davies--Bouldin rank
selects $k{=}\MajSubK$.}
\label{fig:select-k}
\label{fig:majsub-select-k}
\end{figure}

\begin{figure}[ht]
\centering
\begin{minipage}[t]{0.56\linewidth}
\vspace{0pt}
\centering
\includegraphics[width=\linewidth]{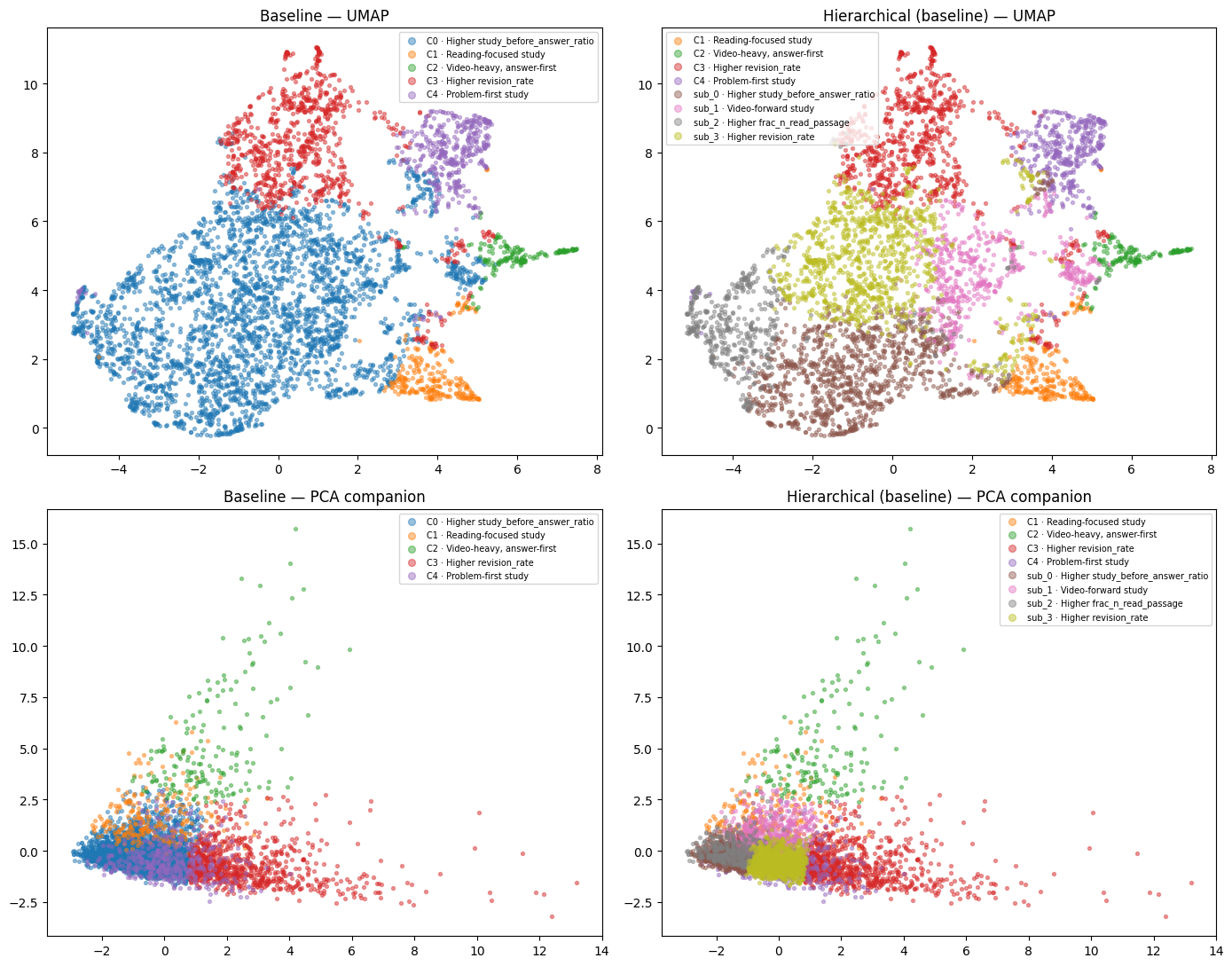}
\end{minipage}\hfill
\begin{minipage}[t]{0.40\linewidth}
\vspace{0pt}%
\centering
\includegraphics[width=\linewidth]{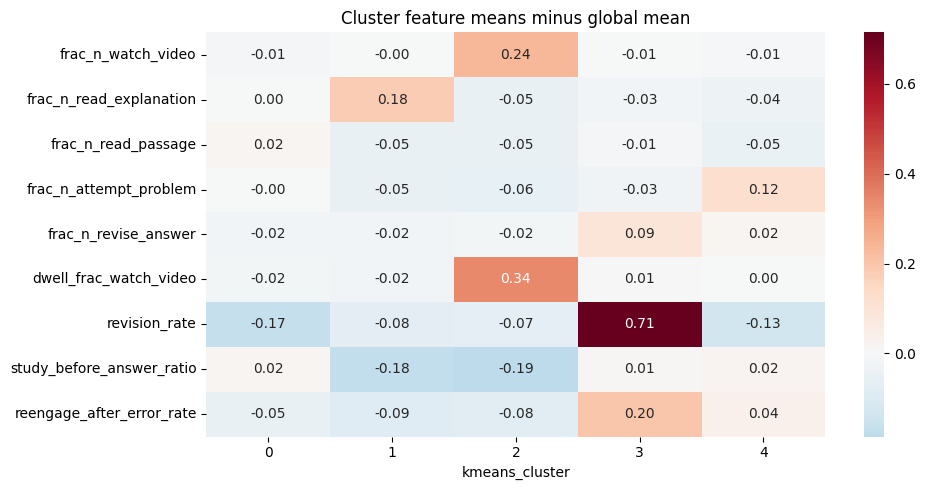}

\vspace{0.6em}
\footnotesize\raggedright
\textbf{(a)} Parents vs.\ hierarchical typology in UMAP (top) and the first
two PCA coordinates (bottom). Left panels: five parent styles. Right panels:
four contrast poles kept fixed while the majority mass is replaced by four
majority styles (prep before answer, video-leaning majority,
passage and attempt, mild revisers). PCA shows poles as elongated tails and
majority styles as a soft decomposition of the dense residual. Neither view
chooses $k$ or assigns labels.

\vspace{0.5em}
\textbf{(b)} Parent-cluster feature contrasts ($\Delta$ vs.\ global mean) that
ground the pole names.
\end{minipage}
\caption{Parent styles and hierarchical typology.
(a)~left: UMAP/PCA views; (b)~right: feature contrasts.}
\label{fig:parent-pca-hier}
\label{fig:parent-char}
\end{figure}

On \Nclustered\ learners, parent $k{=}\LockedK$ recovers \Npoles\ sharp contrast
styles (``poles'') and one large near-average group (the typical majority) with global optimal silhouette and Davis-Bouldin scores (Figure~\ref{fig:select-k}).
Top left of Figure~\ref{fig:parent-pca-hier}(a) shows that cut on UMAP for intuition only, with cluster shares; the actual clustering is in the bottom left PCA space (\NPCA\ components, \PcaVar\ variance). Relative to the
global mean (Figure~\ref{fig:parent-char})(b): a small reading-focused pole
($\sim$5.2\%) with more explanation/passage study and study before answer; a
video-heavy pole ($\sim$3.4\%) with high video share/dwell and less prep or
revision; a larger revision-heavy pole ($\sim$17.9\%) with high revision and
re-engage after error; a problem-first pole ($\sim$8.6\%) with more time on
trying problems than study or preparation; and the typical majority (\MajorityPct), a near-mean residual with
only tiny elevations on study-before-answer/reading. Right side of Figure~\ref{fig:parent-pca-hier}(a) shows the
full hierarchy in both UMAP and the first two PCA coordinates: poles appear as
arms or tails, while the typical majority is a dense cloud that only softens
into four majority styles after a second-stage recluster. 

Reclustering the typical majority (\MajSubN\ users) selects $k{=}\MajSubK$ under
combined silhouette and Davies--Bouldin ranks (right side of Figure~\ref{fig:majsub-select-k}) and yields four softer styles named relative to the \emph{majority} mean
(not the global mean): prep-before-answer (more study-before-answer and
explanation reading), video-leaning majority, passage-and-attempt (more passage
reading and problem-solving attempts), and mild revisers (more revision within
the majority, kept separate from the sharp revision-heavy pole). Parent labels
are highly reproducible under bootstrap resampling (Table~\ref{tab:ari-summary}), even
though within-majority silhouette is modest ($\approx$\MajSubSil). Soft
separation and stable labels can coexist: we treat the hierarchy as a
reproducible descriptive map, not as a set of tight density balls.
To make parent names concrete, we pick one nearest-centroid exemplar per
style (the learner whose feature vector is closest to that cluster's mean) and
plot their first 200 actions over time
(Appendix~\ref{app:exemplars}, Figure~\ref{fig:exemplars}). These timelines show
how action mix and pacing differ across poles, as part of face-validity checks rather than causal evidence.

\subsection{Stability and feature-definition robustness}
Table~\ref{tab:ari-summary} collects every adjusted Rand index (ARI) reported
in the paper so readers need not chase numbers across sections. ARI near $1$
means two partitions agree beyond chance; ARI near $0$ means near-chance
agreement. Bootstrap rows ask whether labels reproduce under resampling;
comparison rows ask whether two different labelings of the same learners
agree.

Bootstrap replicates at fixed $k{=}\LockedK$ show near perfect parent-label agreement. Hierarchical encodings that keep the four poles and replace the typical majority with the
four majority styles remain label-stable, even though within-majority
silhouette is modest ($\approx$\MajSubSil). Soft within-majority silhouette is
therefore compatible with reproducible partitions: stability, not density, is
the gate we use for the reported typology.

\begin{table}[ht]
\caption{Adjusted Rand index (ARI) summary. Bootstrap rows: mean agreement of
resampled labels vs.\ the full-cohort reference (floor $=$ minimum replicate).
Comparison rows: agreement between two labelings of the same learners.}
\label{tab:ari-summary}
\centering
\small
\begin{tabular}{llp{0.38\linewidth}r}
\toprule
Category & Description/comparison & What it asks & ARI \\
\midrule
\multicolumn{4}{l}{\emph{Bootstrap stability (full-timeline typology)}} \\
Bootstrap & Parent $k{=}\LockedK$ & Do parent labels reproduce? &
  \StabARImean\ (floor \StabARImin) \\
Bootstrap & Full hierarchy (\Nhier\ styles) & Do hierarchical labels reproduce? &
  \HierStabARImean \\
Bootstrap & Contrast poles only & Do the \Npoles\ poles reproduce? &
  \PolesStabARImean \\
Bootstrap & Majority styles only & Do the \MajSubK\ majority styles reproduce? &
  \MajSubStabARImean\ (floor \MajSubStabARImin) \\
Bootstrap & Quit-in order cut (exploratory) & Stable with idle-return transitions? &
  \SeqARIbasemean \\
Bootstrap & No-quit order cut (exploratory) & Stable after dropping idle-return? &
  \SeqARImean \\
\midrule
\multicolumn{4}{l}{\emph{Cross-labeling comparisons}} \\
Compare & Early-half vs.\ full-timeline & Same styles, or a fresh early partition? &
  \ValARI \\
Compare & Volume-only vs.\ strategy & Are styles just practice-volume groups? &
  \VolARISameK \\
Compare & KT embedding vs.\ strategy & Does correctness latent match behavior? &
  \KTvsBehARI \\
\bottomrule
\end{tabular}
\end{table}

Ablation studies of feature definitions (Table~\ref{tab:variants})
show that silhouette alone can favor coarser cuts, while Davies--Bouldin often
stays near the locked baseline. Merging reading categories, restricting to
longer sessions, or clustering only on mix/dwell/study-before ratios inflates
silhouette while collapsing poles into a large residual plus a video arm.
Log-count features improve density enough for HDBSCAN to fire (and improve
Davies--Bouldin), but change the naming story. We therefore treat the locked
nine-feature hierarchy as an interpretable and stable baseline, not the
silhouette maximum.

\begin{table}[ht]
\caption{Feature-definition variants at their silhouette-selected $k$.
Silhouette↑ / Davies--Bouldin↓ ($k{=}\LockedK$).}
\label{tab:variants}
\centering
\footnotesize
\setlength{\tabcolsep}{3.5pt}
\begin{tabular}{@{}l c c c p{0.35\linewidth}@{}}
\toprule
Variant & $k$ & Sil. & DB & Structure (role) \\
\midrule
\texttt{locked\_baseline} & 5 & 0.333 & 1.106 &
  Locked 9-feature cut (reference) \\
\texttt{v1\_merged\_read} & 2 & 0.561 & 1.053 &
  Coarser residual + video pole \\
\texttt{v2\_long\_sessions} & 2 & 0.571 & 1.104 &
  Coarser; session-length sensitivity \\
\texttt{v3\_mix\_dwell\_strategy} & 2 & 0.575 & 1.091 &
  Highest silhouette, still coarse \\
\texttt{v4\_log\_counts} & 2 & 0.525 & 0.687 &
  Better DB; HDBSCAN usable; names change \\
\bottomrule
\end{tabular}
\end{table}

\subsection{Exploratory order probes}
\begin{figure}[ht]
\centering
\includegraphics[width=0.85\linewidth]{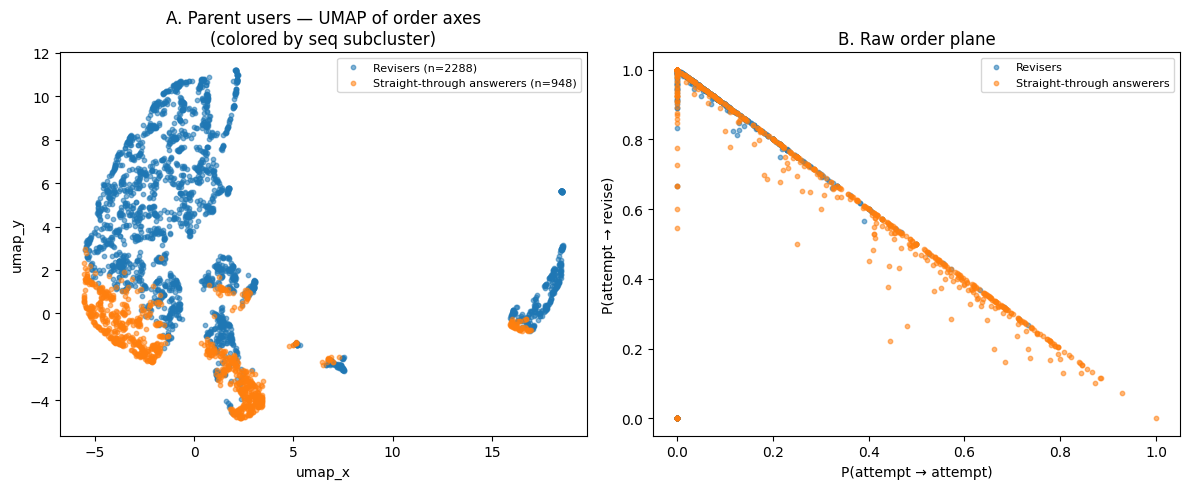}
\caption{Descriptive no-quit order cut inside the typical majority
(exploratory; not part of the locked hierarchy).}
\label{fig:seq-axes}
\end{figure}

As a secondary check, we ask whether \emph{order} of actions (not just how
often each action occurs) splits the typical majority more cleanly. Transition
features that include quit-and-return moves yield a stable binary cut, but after
removing idle-return transitions the same design fails our bootstrap stability
bar with ARI = \SeqARImean (Table~\ref{tab:ari-summary}). We still show the silhouette-selected cut in Figure~\ref{fig:seq-axes}: \SeqReviserPct\ of the majority look like
\emph{revisers} (more attempt-then-revise sequences), and
\SeqAnswerPct\ look like \emph{straight-through answerers} (more
attempt-then-next-attempt, with less revising). We treat that contrast as
descriptive only and do not add it to the locked typology. That cut
separates late persistence ($\eta^2{\approx}$\SeqPersistEta) more than terminal
accuracy ($\eta^2{\approx}$\SeqTermEta), again matching engagement-not-mastery.

\subsection{Early clusters track engagement, not mastery}
\begin{figure}[ht]
\centering
\includegraphics[width=0.95\linewidth]{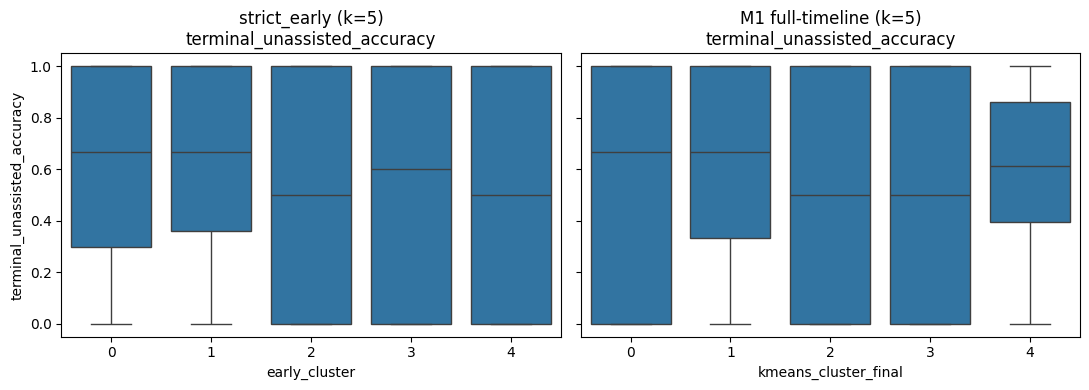}
\caption{Late-period outcomes by early-half cluster membership (anti-leakage
design). Engagement proxies separate more clearly than terminal unassisted
accuracy.}
\label{fig:outcome-box}
\end{figure}

Table~\ref{tab:task10-holm} summarizes the early/late test.
\texttt{strict\_early} recovers \ValEarlyN\ users and agrees only weakly with
full-timeline labels (Table~\ref{tab:ari-summary}), so early cluster ids are not
simply a relabeling of the named poles. After Holm correction, early clusters differ
on late-session persistence
($\eta^2{\approx}$\ValEarlyPersistEta) and late completion
($\eta^2{\approx}$\ValEarlyCompleteEta), but not on terminal unassisted
accuracy
($\eta^2{\approx}$\ValEarlyTermEta; $p_{\mathrm{adj}}{\approx}$\ValEarlyTermPadj).
Figure~\ref{fig:outcome-box} shows the corresponding spreads. In plain terms:
early study-style structure forecasts who keeps practicing and finishing items
later, not who answers later items correctly without help.

Bootstrap bounds on the overall terminal-accuracy gap among scorable labeled
users remain on the order of \MasteryOmnibusPP\ percentage points. Controlling
for total responds leaves the same qualitative ordering. Adjusting for how hard
the late items are mostly shrinks pole-vs-majority gaps toward zero; one
descriptive exception is difficulty-adjusted video-heavy vs.\ majority
(\DiffVideoBPP\,pp; bootstrap 95\% CI \DiffVideoBootCI), on a small video-heavy
cell---not a Holm-controlled mastery claim from the early/late ANOVA.

\begin{table}[ht]
\caption{Holm-adjusted one-way ANOVA of late outcomes under
\texttt{strict\_early} (locked $k{=}\LockedK$).}
\label{tab:task10-holm}
\centering
\begin{tabular}{llrrr}
\toprule
Outcome & Sample \# $n$ & $\eta^2$ & $p_{\mathrm{adj}}$ & Sig. \\
\midrule
persistence\_late\_sessions & 4960 & \ValEarlyPersistEta & $<10^{-100}$ & yes \\
late\_completion\_rate & 4960 & \ValEarlyCompleteEta & $<10^{-20}$ & yes \\
terminal\_unassisted\_accuracy & 2925 & \ValEarlyTermEta & \ValEarlyTermPadj & no \\
\bottomrule
\end{tabular}
\end{table}

\subsection{Volume covaries but does not define styles}
Parent styles differ in how much learners practice (Table~\ref{tab:volume}):
respond-count and action-count ANOVA $\eta^2$ values are \VolRespondEta\ and
$0.025$, while session-count separation is weak ($\eta^2{\approx}0.004$). Mean
responds rise from the reading-focused / typical-majority end of the table to
revision-heavy and problem-first poles. Differing means do not mean the styles
\emph{are} volume groups. Re-clustering the same users on responds, sessions,
and actions alone at $k{=}\LockedK$ barely recovers the strategy labels
(Tables~\ref{tab:ari-summary} and~\ref{tab:volume-baseline}). Among
\OutScorableLabeled\ labeled scorable users, strategy labels separate terminal
accuracy only weakly ($\eta^2{\approx}$\StratMasteryEta) and volume-only labels
even more weakly ($\eta^2{\approx}$\VolMasteryEta). Volume travels with the
typology but does not define it---an important check before using style labels
in adaptive interventions.

\begin{table}[ht]
\caption{Mean volume covariates by full-timeline parent cluster
(\Nclustered\ users). ANOVA $\eta^2$ uncorrected across the three tests.}
\label{tab:volume}
\centering
\small
\begin{tabular}{lrrr}
\toprule
Cluster & Mean responds & Mean actions & Mean sessions \\
\midrule
Typical majority & $\approx$270 & $\approx$1{,}225 & $\approx$16.7 \\
Reading-focused & $\approx$247 & $\approx$1{,}211 & $\approx$19.5 \\
Video-heavy & $\approx$314 & $\approx$1{,}382 & $\approx$21.6 \\
Revision-heavy & $\approx$506 & $\approx$1{,}680 & $\approx$17.6 \\
Problem-first & $\approx$617 & $\approx$1{,}835 & $\approx$19.0 \\
\midrule
ANOVA $\eta^2$ & \VolRespondEta & 0.025 & 0.004 \\
\bottomrule
\end{tabular}
\end{table}

\begin{table}[ht]
\caption{Activity-only baseline vs.\ strategy labels ($k{=}\LockedK$;
$n{=}$\OutScorableLabeled\ scorable labeled users for mastery columns).
ARI also appears in Table~\ref{tab:ari-summary}.}
\label{tab:volume-baseline}
\centering
\begin{tabular}{lrrr}
\toprule
Labeling & ARI vs.\ strategy & Terminal $\eta^2$ & Omnibus range (pp) \\
\midrule
Strategy & --- & \StratMasteryEta & \MasteryOmnibusPP \\
Volume-only & \VolARISameK & \VolMasteryEta & 6.27 \\
\bottomrule
\end{tabular}
\end{table}

\subsection{Supervised KT probe}
EdNet questions carry both a coarse \texttt{part} label (TOEIC exam section
$1$--$7$) and finer expert \texttt{tags}. We use \texttt{part} only: the probe
is a correctness-supervised contrast to behavior clusters, not a KT bake-off, so
seven section-level knowledge components are enough, and they match our
difficulty baseline. On \KTnUsers\ learners with those \KTnSkills\ section
labels, SAKT reaches validation AUC~\KTauc. We compare it to a
\emph{part-difficulty prior} on the same attempts: if the next item is in
section~$p$, predict using only the cohort's success rate on section~$p$ (the
same probability for every learner; no personal history). That naive baseline
reaches AUC~\KTbase; SAKT's lift is \KTlift\ (Table~\ref{tab:kt-summary}). A
user-level bootstrap 95\% CI on the AUC difference is \KTliftCI\ and excludes
zero: answer history adds a small, reliable signal beyond ``this section is
usually this hard.'' Absolute AUC remains far below full-EdNet KT benchmarks
($\sim$0.78), as expected on a $\sim$5k cohort: we emphasize the lift, not a
leaderboard comparison.

The $64$-d SAKT embeddings largely re-encode ability (PC1 variance share
\KTpcOnevar; corr(PC1, mean correctness)$\,{=}\,$\KTpcOner;
$R^2{\approx}$\KTabilityR). A silhouette-selected binary cut of those
embeddings is bootstrap-stable but diffuse (HDBSCAN noise $88$--$100$\%). It
barely agrees with the locked behavioral typology
(Table~\ref{tab:ari-summary}) yet separates terminal accuracy
($\eta^2{\approx}$\KTtermEta), as expected for a model trained on correctness,
and not evidence of a dense behavioral mastery typology. Behavior-only
clustering and correctness-supervised embeddings answer different questions, and neither substitutes for the other on this cohort.

\begin{table}[ht]
\caption{Supervised KT probe summary. Items are labeled by TOEIC
\texttt{part} (sections $1$--$7$). The part-difficulty prior is a naive
baseline: for section~$p$, use the cohort's success rate on~$p$ for every
learner (no personal history). Lift is
$\mathrm{AUC}(\mathrm{SAKT})-\mathrm{AUC}(\mathrm{prior})$.}
\label{tab:kt-summary}
\centering
\begin{tabular}{ll}
\toprule
Quantity & Value \\
\midrule
Learners with KT sequences & \KTnUsers \\
TOEIC exam sections (\texttt{part}) & \KTnSkills\ ($1$--$7$) \\
SAKT validation AUC & \KTauc \\
Part-difficulty prior AUC
  (cohort success rate on section) & \KTbase \\
AUC lift over prior (95\% CI) & \KTlift\ (\KTliftCI) \\
KT-cluster vs.\ behavior ARI & \KTvsBehARI \\
Terminal-accuracy $\eta^2$ (KT clusters) & \KTtermEta \\
\bottomrule
\end{tabular}
\end{table}

\section{Discussion}

A common inference in behavior-based learner modeling is that a stable,
interpretable, well-separated cluster must therefore predict learning. Our
results caution against that leap. We recover a stable hierarchy of study styles
that is easy to narrate, and that feature ablations show is not an artifact of
chasing maximum silhouette, yet clusters built from early practice forecast
later engagement rather than later unassisted accuracy. That split matches the
broader engagement--achievement literature
\cite{fredricks2004school,henrie2015measuring}: styles can guide \emph{how} to
support consistent practice without implying a mastery diagnosis.  

Furthermore, parent styles
barely differ in mean empirical item difficulty, and difficulty-adjusted OLS
does not overturn the overall mastery null, though the small video-heavy pole
remains a descriptive caveat (\DiffVideoBPP\,pp vs.\ majority;
bootstrap CI \DiffVideoBootCI). Our claim is therefore precise: behavior-only
profiles do not predict later \emph{in-app accuracy} in general on this cohort.
A link (or null) to external learning gains would require independent outcome
labels.

\textbf{Implications for analytics practice.} Policies keyed only on labels such
as ``revision-heavy'' or ``video-heavy'' will partly track who practiced more
and who stays active later, not who knows more. Correctness-aware models
(including knowledge tracing) can recover ability structure, but that structure
is nearly independent of behavior-only profiles here
(Table~\ref{tab:ari-summary}).
Practitioners should pair style dashboards with independent mastery evidence
(item correctness, external assessments) rather than treat cluster membership
as a knowledge proxy. If the institutional goal is sustained practice, early
behavioral structure \emph{is} informative: persistence is the signal, not
terminal accuracy.

\textbf{Theoretical reading.} From an item-response perspective
\cite{rasch1960,embretson2000irt}, models trained on correctness are expected
to recover an ability-like direction; clustering those latents can separate
mastery without implying a behavioral typology. Our SAKT probe matches that
expectation. Conversely, our study-strategy features deliberately omit
correctness, so their clusters need not and here do not align with ability.
The evaluative contribution is to document that separation when early features
and late outcomes are kept apart, rather than assume educational meaning from
silhouette alone.

\section{Limitations}
We state how far the conclusions should be taken. These bounds are concrete
measurement and design limits, not generic caveats.

\emph{In-app proxies, not external outcomes.}
All outcomes are computed from the same EdNet-KT3 interaction logs that produce
the behavioral features. ``Mastery'' here means accuracy on later
\emph{in-app} questions, specifically terminal unassisted first-attempts in
the late respond-ranked half, not a post-test, course grade, certification, or
retention label (EdNet-KT3 does not ship those). In-app accuracy remains
confoundable by guessing and by which items a learner reaches.

\emph{Sampling frame and representation scope.}
Results describe active learners ($\geq$50 \texttt{respond}s) on one
TOEIC-oriented platform. They do not automatically generalize to less active
users, K--12 settings, other subjects, or other ITS products. Of \Nfeatures\
users in the feature table, \Ndropped\ are dropped before PCA because at least
one clustering feature is non-finite (almost always an undefined dwell fraction
when the user has no timed study dwell); we do not impute those rows, leaving
\Nclustered\ labeled learners. The reported typology is the hierarchical
\Nhier-style encoding; the typical majority's second-stage silhouette is modest
($\approx$\MajSubSil) even though hierarchical label agreement is high
(\HierStabARI; majority styles \MajSubStabARI). Soft silhouette does not
overturn label stability, but feature-definition variants can prefer coarser,
higher-silhouette partitions. The parent bootstrap we report uses \StabNrep\
replicates (config default is 10); expanding the replicate count is unlikely to
change the qualitative conclusion given the observed ARI floors
(\StabARImin\ parent; \MajSubStabARImin\ for majority styles). Hand-engineered
flat and Markov features may miss topic-level or multi-scale sequential
structure; SAKT is a single, relatively small architecture used as a supervised
latent probe rather than a KT bake-off. Absolute SAKT AUC (\KTauc) is below
full-EdNet benchmarks because we train on a $\sim$5k active cohort; we
therefore emphasize lift over the part-difficulty prior (\KTlift; CI
\KTliftCI)---cohort success rate on the item's TOEIC section, with no personal
history---not a state-of-the-art comparison. KT embeddings use the
early-stopped validation checkpoint and are a descriptive supervised probe, not
a pure hold-out embedding study.

\emph{Respond-rank 50/50 split and unequal late volume.}
The early/late cut is respond-count proportional at a default 50/50 fraction,
not calendar or checkpoint based. Equalizing experience \emph{share} still
leaves unequal late \emph{volume}: the late half can be on the order of 10
responds for one learner and 1{,}000 for another, so the reliability of
terminal accuracy is heteroskedastic. Users with few late or few unassisted
terminal items contribute noisier mastery estimates, and only
\OutScorable\ of \Nusers\ ($\sim$59\%) are scorable on that proxy at all.
Labeled-scorable mastery analyses further intersect full-timeline labels
(\OutScorableLabeled). The mastery claim is therefore about the
\emph{detectable mean cluster gap} (with bootstrap CIs among scorable users),
not that every learner has an equally precise mastery measurement. Session
features also discard fine-grained within-action timing. Alternate 40/60 and
60/40 splits leave the terminal gap qualitatively unchanged---evidence that the
mastery null is not an artifact of the exact halfway cut, but only a weak
sensitivity check of the early/late boundary for non-terminal outcomes. We did
not adopt a criteria-based cut (fixed early respond floor, content unit, or
correctness gate) because EdNet-KT3 lacks shared course checkpoints and
outcome-tied gates risk soft leakage into the split itself.

\emph{Full-timeline names vs.\ early association labels.}
Profile names and stability ARIs describe the full-timeline lock
(Table~\ref{tab:ari-summary}). The \texttt{strict\_early} partition uses
the same fixed $k{=}\LockedK$ but is an independent fit on early features; its
weak agreement with full-timeline labels means it should not be narrated with
the named pole/sub labels. We also
do not re-select $k$ on the early half, so early structure is a locked-cardinality
transfer check rather than a fresh model selection. Majority style names such as
``prep-before-answer'' are relative to the typical-majority mean, slight
elevations within a soft residual, rather than sharp, globally unique strategies.

\emph{Secondary audits: sequential actions.}
Transition features that include leave-and-come-back (idle\_return) events yield a stable binary cut on the typical majority, but after removing those events the cut is no longer bootstrap-stable (Table~\ref{tab:ari-summary}). We therefore treat the reviser (first attempt on a problem then come back) vs.\ straight-through-answerer (keep answering new questions) contrast as descriptive only and do not add it to the reported \Nhier-style typology.
Qualitative exemplars (Appendix~\ref{app:exemplars}) cannot substitute for
outcome validation.

\emph{Actionability.}
A stable typology that tracks engagement does not by itself prescribe
institutional interventions (who to message, how to redesign content).
Translating descriptive profiles into educator-facing or student-facing supports
would require co-design with practitioners and outcome definitions that match
institutional goals. Stronger negative or positive claims would need
multi-platform replication and external criteria.

\section{Conclusion}

Unsupervised study-strategy clustering on EdNet-KT3 yields a stable hierarchy of
\Nhier\ styles: sharp contrast poles at the first cut plus softer styles inside
the typical majority. When clusters are built from early practice only, they
forecast later engagement, especially whether learners stay active across late
sessions, rather than later unassisted accuracy. Practice volume travels with parent
styles but does not define them. A supervised knowledge-tracing probe on TOEIC
exam sections beats a section-difficulty baseline only modestly and recovers
mastery as an ability gradient nearly independent of behavior styles. For
learning analytics aiming to help learners thrive, behavioral clusters remain
valuable for describing study practices and supporting engagement, but claims
about knowledge gains need correctness-aware evidence beyond behavior-only
labels.

\textbf{Toward helping learners thrive.} LAK27's theme asks analytics not only
to describe or predict, but to support meaningful educational outcomes. Our
results clarify a precondition for that ambition in ITS settings: behavior-only
profiles are actionable for engagement support, yet insufficient alone as
mastery signals. Future work should test two practical next steps: (1)~nudge
learners who show early signs of dropping practice (not because they match a
named style such as revision-heavy) and check whether that improves how
consistently they keep studying; and (2)~show study-style dashboards together
with separate mastery evidence (knowledge-tracing estimates or assessments) and
test whether teachers or systems decide better with both signals than with
styles alone or mastery alone.

\textbf{Ethics and stakeholder use.} EdNet is a public research corpus; our
analyses use aggregated behavioral features without attempting to re-identify
learners. In a live ITS, treat style labels as descriptions of how learners
practice and as cues for engagement support, rather than as measures of knowledge.
Because behavior styles did not reliably predict later unassisted accuracy
here, do not use them for high-stakes decisions such as ranking students,
flagging ``weak'' learners, or withholding opportunities based on membership in
a pole (e.g., revision-heavy).

\textbf{Reproducibility.} Paths, thresholds, and seeds live in a
configuration file; each pipeline stage writes run metadata. Analysis code and
notebooks for the locked results are available from the author upon reasonable
request (and will be linked in a camera-ready version if applicable).

\section*{Acknowledgments} 
\addcontentsline{toc}{section}{Acknowledgments} 
The authors thank course staff and peers for feedback on earlier drafts of this
work.

\section*{Declaration of Conflicting Interest}
\addcontentsline{toc}{section}{Declaration of Conflicting Interest}
The authors declared no potential conflicts of interest with respect to the
research, authorship, and/or publication of this article.

\section*{Funding}
\addcontentsline{toc}{section}{Funding}
This research received no specific grant from any funding agency in the public,
commercial, or not-for-profit sectors.

\phantomsection
\bibliography{references_project}

\section*{Appendix}
\appendix
\subsection{Exemplar parent-style timelines}
\label{app:exemplars}
\begin{figure}[ht]
\centering
\includegraphics[width=0.92\linewidth]{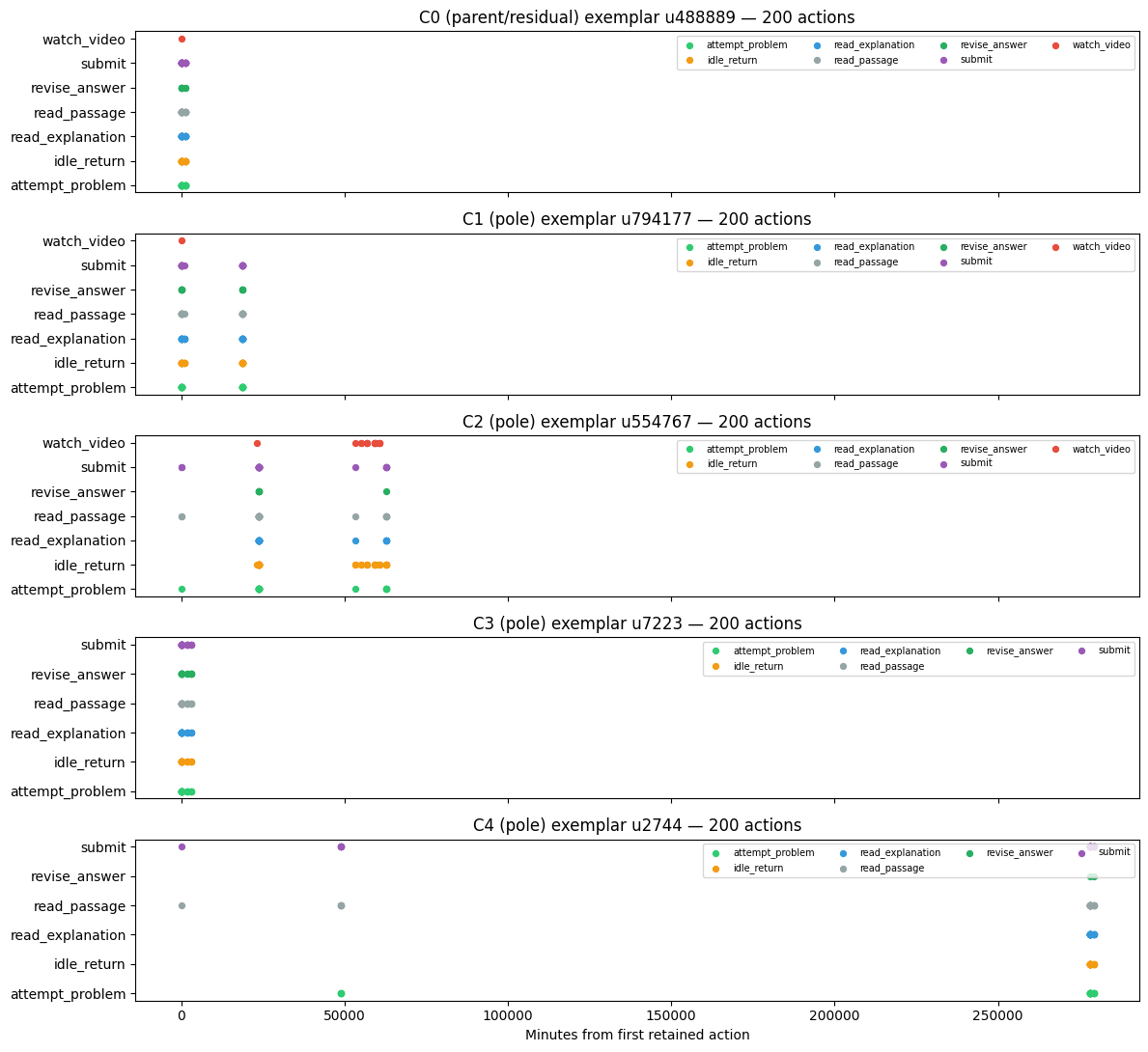}
\caption{Nearest-centroid exemplar action timelines for parent styles (first
200 actions).}
\label{fig:exemplars}
\end{figure}

Figure~\ref{fig:exemplars} shows nearest-centroid exemplars (first 200 actions)
for each parent style. Typical-majority and revision-heavy exemplars concentrate
activity in a short early burst; the reading-focused exemplar shows a second
session after a long idle gap; the video-heavy exemplar mixes video with
repeated returns; the problem-first exemplar stretches sparse problem-solving
across a long calendar span. Temporal density and action mix differ sharply
even when late mastery does not; these timelines are illustrative face-validity
checks, not causal evidence.

\end{document}